\documentclass{article}
\usepackage{microtype}
\usepackage{graphicx}
\usepackage{subfigure}
\usepackage{booktabs} 

\usepackage[normalem]{ulem}

\usepackage{url}
\usepackage{hyperref}
\usepackage[hyphenbreaks]{breakurl}

\usepackage{xcolor}
\usepackage{changes}
\newcommand{\defineReviewer}[2]{
    \definecolor{color-#1}{rgb}{#2}
    \definechangesauthor[name={#1}, color=color-#1]{#1}
    \expandafter\newcommand\csname #1Add\endcsname[1]{\added[id=#1]{##1}}
    \expandafter\newcommand\csname #1Rem\endcsname[1]{\deleted[id=#1]{##1}}
    \expandafter\newcommand\csname #1Rep\endcsname[2]{\replaced[id=#1]{##2}{##1}}
}

\defineReviewer{G}{1, 0.2, 0.2}
\defineReviewer{J}{0.2, 0.2, 1}

\usepackage[acronym]{glossaries}
\newacronym{agi}{AGI}{Artificial General Intelligence}
\newacronym{ai}{AI}{Artificial Intelligence}
\newacronym{llm}{LLM}{Large Language Model}
\newacronym{pov}{PoV}{Point of View}
\newacronym{eai}{E-AI}{Embodied AI}
\newacronym{iai}{I-AI}{Internet AI}
\newacronym{tame}{TAME}{Technological Approach to Mind Everywhere}
\newacronym{rl}{RL}{Reinforcement Learning}
\newacronym{rlhf}{RLHF}{Reinforcement Learning from Human Feedback}
\newacronym{nn}{NN}{neural network}
\newacronym{rag}{RAG}{Retrieval-Augmented Generation}

\usepackage[accepted]{icml2024}
\icmltitlerunning{A Call for Embodied AI}

\begin{document}
\allowdisplaybreaks

\twocolumn[
\icmltitle{A Call for Embodied AI}


\begin{icmlauthorlist}
\icmlauthor{Giuseppe Paolo}{to}
\icmlauthor{Jonas Gonzalez-Billandon}{lon}
\icmlauthor{Bal\'{a}zs K\'{e}gl}{to}
\end{icmlauthorlist}

\icmlaffiliation{to}{Noah's Ark Lab, Huawei Technologies France, Paris, France}
\icmlaffiliation{lon}{London Research Center, London, UK}

\icmlcorrespondingauthor{Giuseppe Paolo}{giuseppe.paolo@huawei.com}
\icmlkeywords{Machine Learning, ICML, Position Paper, Embodied AI}

\vskip 0.3in
]
\printAffiliationsAndNotice{}

\begin{abstract}
We propose Embodied AI (E-AI) as the next fundamental step in the pursuit of Artificial General Intelligence (AGI), juxtaposing it against current AI advancements, particularly Large Language Models (LLMs). 
We traverse the evolution of the embodiment concept across diverse fields (philosophy, psychology, neuroscience, and robotics) to highlight how E-AI distinguishes itself from the classical paradigm of static learning. 
By broadening the scope of E-AI, we introduce a theoretical framework based on cognitive architectures, emphasizing perception, action, memory, and learning as essential components of an embodied agent. 
This framework is aligned with Friston’s active inference principle, offering a comprehensive approach to E-AI development. 
Despite the progress made in the field of AI, substantial challenges, such as the formulation of a novel AI learning theory and the innovation of advanced hardware, persist. 
Our discussion lays down a foundational guideline for future E-AI research. 
Highlighting the importance of creating E-AI agents capable of seamless communication, collaboration, and coexistence with humans and other intelligent entities within real-world environments, we aim to steer the AI community towards addressing the multifaceted challenges and seizing the opportunities that lie ahead in the quest for AGI. 
\vspace{-6pt}
\end{abstract}

\section{Introduction}
\vspace{-4pt}

Over recent years, the field of artificial intelligence (AI) has experienced a significant surge, leading to substantial breakthroughs in areas ranging from computer vision (CV) and natural language processing (NLP) to neuroscience. This journey through AI's development has been marked by a series of significant triumphs interspersed with setbacks, including the well-documented AI winter of the mid-1980s. The ambitious goal that has propelled AI research forward from the beginning was to create intelligence that either parallels or exceeds human abilities.
This quest for superhuman intelligence, commonly termed Artificial General Intelligence (AGI), has been seen differently by experts across different disciplines, yet it broadly refers to the ability of a system to understand, learn, and apply knowledge in a wide array of tasks and contexts, mirroring the cognitive flexibility of humans and animals. 

The remarkable progress in AI over the past decade can largely be attributed to three pivotal developments: i) advancements in deep learning algorithms, ii) the advent of powerful new hardware, and iii) the availability of extensive datasets for training. A prime illustration of this advancement is the creation of Large Language Models (LLMs) like OpenAI's GPT-4 \citep{achiam2023gpt} and Google's Gemini \citep{team2023gemini}. 
The surprising abilities of these LLMs have sparked discussions within the AI community, with some pondering whether these models have already achieved nascent forms of AGI. Foundation models (large networks with billions of parameters trained on massive datasets) have found success in varied fields, ranging from predicting 3D protein structures \cite{cramer2021alphafold2} and robotic control \cite{rt22023arxiv}, to generating images and audio \cite{ramesh2022hierarchical, radford2022robust}. 
This breadth of achievement supports the hypothesis that continued scaling and refinement of foundation models could be a viable path toward realizing AGI.

In our paper, we argue that despite the significant advances made by current AI technologies, they represent only the initial steps towards truly intelligent agents. Despite their impressive capabilities, these large networks are static and unable to evolve with time and experience.
They leverage large datasets and cutting-edge hardware for scaling, but they lack the ability to properly \emph{care} about the truth \cite{Vervaeke2024}, which in turn makes it impossible to dynamically adjust their knowledge and actively search for \emph{valuable} new information.
The two primary manifestations of this fundamental shortfall are i) the difficulty in effectively aligning LLMs \cite{ouyang2022training}, and ii) their propensity to generate plausible but inaccurate information, a phenomenon known as confabulation \cite{huang2023survey, wei2023chainofthought}.
Current strategies to mitigate these issues, such as post-processing, fine-tuning, prompt engineering, and incorporating human feedback, are undeniably valuable. 
However, we argue that these methods address only the superficial aspects of the problem and fall short in dealing with the core issue at play: the inherent lack of a deeper, grounded sense of \emph{care} in LLMs.

Pursuing the development of AGI, we draw upon the insights of \citet{Vervaeke2024} to advocate for designing AI agents that are bound to, observe, interact with, and learn from the real world (including humans) in a continuous and dynamic manner. These \gls{eai} agents ought to prioritize their continued existence and our bindings to them, thereby learning the \emph{value} of truth. They should also be capable of adapting to environmental changes and evolving without human intervention.

While Large Language Models play a significant role in the development of AI systems, they fall short of capturing the essence of what constitutes an intelligent agent. Notably, intelligent beings, whether humans or animals, are characterized by three fundamental components: the mind, perception, and action capabilities \cite{kirchhoff2018markov}. LLMs, or more broadly, foundation models, may be likened to an aspect of the mind's reasoning function \cite{xi2023rise}. Yet, the perceptive and action-oriented dimensions of intelligence, along with the pivotal ability to dynamically revise beliefs and knowledge based on experiences, remain unaddressed.
Autoregressive LLMs are not designed to understand the causal relationships between events, but rather to identify proximate context and correlations within sequences \cite{bariah2023ai}. 
In contrast, a fully embodied agent should have the ability to grasp the causality underlying events and actions within its environment, be it digital or physical. 
By comprehending these causal relationships, such an agent can make informed decisions that consider both the anticipated outcomes and the reasons behind those outcomes.

In short, \textbf{this paper argues that the necessary next step in our pursuit for truly intelligent and general AIs is the development and study of Embodied AI.
We propose that current LLM-based foundation models could lay the groundwork for designing these agents, but are just one component of a truly embodied agent. }
This approach is akin to how neonates come into the world equipped with inherent priors to successfully adapt to the world \cite{reynolds2018development}.

In the next section of this paper, we will define the concept of embodiment and what we mean by \gls{eai} in Sec.~\ref{sec:definition_eai}, analyzing the literature and various scientific and philosophical currents.
In Sec.~\ref{sec:Why_Embodiement}, we discuss why we believe this is a necessary step towards \gls{agi}. 
In Sec.~\ref{sec:theoretical_framework} we analyze the main components of a truly embodied agent, and we discuss the major challenges to achieving this ambitious goal in Sec.~\ref{sec:challenges}. 
Our motivation behind the need to develop \gls{eai} and its fundamental role in our path towards \gls{agi} is proposed throughout. 
Finally, Sec.~\ref{sec:conclusion} provides a short recap of our proposition.
\vspace{-4pt}

\section{What is embodied AI}
\vspace{-4pt}
\label{sec:definition_eai}

\glsfirst{eai} is s sub-field of AI, focusing on agents that interact with their physical environment, emphasizing sensorimotor coupling and situated intelligence. As opposed to mere passive observing, \gls{eai} agents act on their environment and learn from the reaction. \gls{eai} is deeply rooted in \emph{embodied cognition} \cite{Shapiro2011,mcnerney2011}, a perspective in philosophy and cognitive science that posits a profound coupling between the mind and the body. This idea, challenging Cartesian dualism — the historically dominant view that distinctly separates the mind from the body \cite{descartes2012discourse} — emerged in the early 20th century. Pioneers like \citet{lakoff1979metaphors, lakoff1999philosophy} have significantly contributed to this paradigm by proposing that reason is not based on abstract laws but is  grounded in bodily experiences.
Embodied cognition forms a critical part of the 4E cognitive science framework \cite{Varela1991, Clark1997, Clark1998}, encompassing embodied, enactive, embedded, and extended aspects of cognition. Within \gls{eai}, the focus is predominantly on implementing the `embodied' and `enactive' aspects, while the `embedded' and `extended' components are more pertinent to situating AI in a social context and as an augmentation of human (individual or collective) cognition.

In AI, initial explorations into embodiment emerged in the 1980s, driven by a growing recognition of the inherent limitations in disembodied agents. 
These limitations were primarily attributed to the absence of rich, high-bandwidth interactions with the environment \citep{pfeifer2004embodied, Pfeifer2006}. 
An early advocate for this paradigm shift was \citet{brooks1991intelligence}, who built walking robots simulating insect-like locomotion.
Simultaneously, the field of computer vision was undergoing its own transformation. 
Researchers and practitioners were increasingly focusing on enabling agents to interact with their surroundings. 
This emphasis on interaction led to a concentration on the perceptual elements of embodiment, particularly from a first-person point of view (POV) \citep{shapiro2021computer}. 
This approach aligns with the concept of visual exploration and navigation \cite{ramakrishnan2021exploration}, where an agent acquires information about a 3D environment through movement and sensory perception, thereby continuously refining its model of the environment \cite{anderson2018vision, chen2019learning}. Such exploration techniques empower an agent to discover objects and understand their permanence.
As a result of these developments, many contemporary benchmarks in \gls{eai} have emerged predominantly from the domains of vision and robotics \cite{duan2022survey}, reflecting the integral role these disciplines have played in advancing the field.

That said, the broader definition of \gls{eai} does not require vision. Sensorimotor coupling may be implemented using \emph{any} physical sense \citep{Pfeifer2006}. In the living world, many organisms survive and thrive without vision, using, for example, chemical or electric sensing \citep{Bargmann2006}. \citet{levin2022technological}'s \gls{tame} framework further explores this idea, suggesting that cognition emerges from the collective intelligence of cell groups, they themselves deeply embodied within their environment (the body they comprise). This framework challenges traditional Cartesian dualism, embedding cognition within the physical and biological makeup of an organism.
In the \gls{tame} perspective, cognition is not just an attribute of higher-order organisms; it extends throughout the ontological hierarchy of living beings, from individual cells, through tissues and organs, to complex organisms. Each agent demonstrates cognitive capabilities that are inherently connected to its physical structure and the environmental interactions at its proper level. This broadened view of cognition and embodiment goes beyond the conventional focus on vision in robotics and computer vision. It posits that any entity capable of perceiving, interacting with, and learning from its environment, thereby adapting to it and influencing it, qualifies as embodied.
A technological instantiation of this concept is an intelligent router in a telecommunication network. This device `lives' in a realm dominated by electromagnetic sensing. It continuously learns from and adapts to the network traffic, effectively mapping and managing the flow of information. This example underscores the potential of applying the principles of \gls{eai} beyond the traditional domains, embracing a more inclusive and diverse understanding of intelligence and embodiment.

This broadening of the notion of \gls{eai} raises the question: \emph{how close current commercial AI tools are to embodiment?} Here we examine two such tools: Large Language Models \citep{brown2020language,devlin2019bert} and Social Media Content AI Recommendation Systems (SMAI) \citep{Bakshy2015, Covington2016, Eirinaki2018}.

LLMs operate within a linguistic-symbolic domain, representing textual information and generating new text by completing prompts. Their foundational training is essentially static, relying on datasets meticulously compiled and curated by teams of AI engineers. Their goal is supervised: to generate likely tokens following a context.
Their secondary training (fine-tuning) may involve both interactions with their symbolic environment (human users) and goals to reach (satisfy their human users), but these interactions are presents some limits due to 
both technical (e.g., catastrophic forgetting \citep{Kirkpatrick2017, Parisi2019}) and business (e.g., managing individuated LLMs \citep{Strubell2019, Kaplan2020}) reasons.
Looking ahead, we anticipate advancements that might address these limitations, potentially leading to the emergence of ``personal assistant'' LLMs. These would represent a form of embodied agents within a symbolic realm. However, at present, LLMs largely resemble static \emph{\gls{iai}} \cite{duan2022survey}, differing significantly from the dynamic, interactive nature characteristic of \gls{eai}.

It is intriguing that despite the growing concerns about the risks and alignment challenges of LLMs highlighted in recent research \citep{Bender2021}, SMAIs have attracted comparatively less scrutiny \citep{Huszar2022, Ribeiro2020}.
This is noteworthy considering SMAIs have been around for a longer time and their influence on society is both wider and more profound. 
We propose that their widespread acceptance and their more integrated, less intrusive presence in our lives are due to their closer alignment with the principles of embodiment, in contrast to LLMs.

What do we mean by SMAIs being closer to embodiment? Firstly, SMAIs are driven by clear objectives: to captivate our attention and maximize our engagement with their respective platforms \citep{Bozdag2013, Bodo2021}. These goals are fundamentally linked to the business models of these platforms, which revolve around advertising. The specifics of these ``engagement'' objectives are typically proprietary, forming the core of the competitive advantage of these platforms. Although these goals are initially human-designed and not intrinsically generated \citep{Covington2016}, they are subject to evolutionary pressure and adaptation, and thus they are tied to the existence and the survival of the SMAI.
Secondly, SMAIs learn almost entirely from the data they collect by interacting with us. 
This leads to a high level of individuation (adapting to our individual preferences \citep{Nguyen2014}), and notions of exploration (offering us content not so much to satisfy us but for the sake of learning what we like). 
This creates a user experience that, when well-executed, resembles interaction with a considerate friend, who wants our best, who connects us to things we like, and who wants to understand us better. 
The flip side, however, is the potential for these systems to morph into mechanisms that perpetuate addictive behaviors or harmful content \citep{Schull2012}. 
Nevertheless, since SMAIs connect and adapt to us in a more intuitive and deeper manner than LLMs, we often feel a greater sense of control over our interactions with these systems (by, for example, consciously not clicking on content that we know we do not want to see in the long run). 
This control, albeit limited, is reminiscent of \emph{persuasion} more than mechanical manipulation, aligning with how we interact with other sentient beings rather than machines. 
This type of relationship with AI systems is a fundamental aspect of \citet{levin2022technological}'s TAME proposal .
Our stance on \gls{eai} suggests that, while systems akin to SMAIs pose greater risks due to their seamless integration into our social fabric, they also present more natural opportunities for alignment with our values. This alignment process is procedural, perspectival, and evolutionary in nature \citep{Vervaeke2012,Vervaeke2024}, contrasting with the primarily propositional approaches being applied to LLMs \citep{shen2023large}.

We posit that the potential for more effective and naturally aligned AI systems is, alone, a compelling reason to prioritizing \gls{eai} in the broader AI research agenda.

In the forthcoming section, we further explore the pivotal role that well-executed implementations of \gls{eai} could play in the quest for AGI.
\vspace{-4pt}

\section{Why embodiement?}
\label{sec:Why_Embodiement}
\vspace{-4pt}

In the previous section, we examined how contemporary theories of embodiment, particularly the TAME framework \citep{levin2022technological}, challenge the long-standing Cartesian dualism which posits a distinct separation between mind and body \citep{descartes2012discourse}. 
This philosophical stance has significantly influenced the development of current generative AI models, such as LLMs, which primarily rely on static data and lack interaction with the physical or even the symbolic world. It is a prevalent belief that simply scaling up such models, in terms of data volume and computational power, could lead to AGI. We contest this view. We propose that true understanding, not only propositional truth but also the \emph{value} of propositions that guide us how to \emph{act}, is achievable only through \gls{eai} agents that live in the world and learn of it by interacting with it.

The significance of embodiment in cognitive development was  demonstrated by \citet{held1963movement}'s carousel experiment with kittens. In this study, one kitten could actively interact with and control a carousel, while the other could only observe it passively. 
Despite both kittens receiving identical visual input, the one engaged in active interaction exhibited normal visual development, unlike its passively observing counterpart. 
This seminal experiment underscores the vital role of embodied interaction in shaping cognitive abilities \cite{shenavarmasouleh2022embodied}. It also reinforces the observation that all known forms of intelligence, including human intelligence, are inherently embodied \citep{smith2005development}, suggesting that embodiment serves as a solid foundation for cognitive learning and development. 
Current AI learns in a very different way from humans. 
We humans learn by seeing, moving, interacting with the world and speaking with others. 
We also learn by collecting sequential experiences, not by passive observation of shuffled and randomized, even if carefully selected, data \cite{smith2005development, westho2020social}. 
We advocate for an approach where insights from cognitive science and developmental psychology inform the design of AI systems. Such systems should be designed to learn through active interaction with their surroundings, mirroring the embodied learning processes fundamental to human cognition.

Even advocates of static learning concede that multimodal learning is the next milestone towards AGI \citep{fei2022towards, parcalabescu2021multimodality}. In \gls{iai}, multimodal data needs to be collected and connected painstakingly. In contrast, \gls{eai} agents, when equipped with multimodal sensors, will inherently collect and correlate multi-modal data by mere co-occurrence. For instance, robots will see (CV), communicate (NLP), reason (general intelligence), navigate and interact with their environment (planning and RL), all simultaneously \cite{shenavarmasouleh2022embodied}. Intelligent routers will observe requests and traffic (sensing), communicate with other routers, human engineers, absorb news about their surroundings (NLP), reason (general intelligence), and control the traffic (control and RL). Despite the impressive progress in these domains, much of it has relied on the external collection and curation of vast datasets for algorithmic training. This approach has significant drawbacks: i) the collection and preparation of data demands substantial investments; ii) this data can contain biases that are hard to detect and rectify \citep{li2020deeper, balayn2021managing, verma2021removing}. 
The issue of biases is particularly pertinent in discussions on AI alignment \citep{shen2023large, ji2023ai}. 
Efforts to align AI through rule-based and procedural methods (such as RLHF \cite{lambert2022illustrating}) often struggle, producing systems that feel mechanistic and ``dumb'', rather than an agent which seamlessly \emph{acts} according to values compatible with our society.

An embodied agent, designed to interact with and learn from its environment, fundamentally changes the traditional approach to data collection and curation in AI development. By being inherently integrated with its physical and social contexts, such an agent bypasses the labor-intensive processes previously required. This shift not only simplifies the challenge of aligning AI with human values but also enhances the agent's learning efficiency by utilizing the unique features of its environment. As a result, the focus in AI development transitions from data to \emph{simulators}. 
These simulators serve a dual purpose: they are both training grounds for \gls{eai} and platforms for testing and refining concepts and algorithms \citep{duan2022survey}. 
Moreover, the process of aligning these agents with human values becomes more intuitive as it involves defining goals reflective of those values. 
This approach does not claim to fully resolve the alignment challenge, as \gls{eai} systems will still necessitate oversight and guidelines to avert unwanted behaviors. 
However, the alignment process becomes inherently more natural. 
Adjusting and defining \emph{goals} is a more straightforward task than the extensive editing and curating of \emph{data}. 
This methodology draws upon our inherent, non-propositional understanding and instincts about aligning embodied intelligences—whether it is guiding our own actions, nurturing children, or training pets.

Another important characteristic of \gls{eai}, stemming from the coupling between the agent and its environment, is the agent's capacity for ongoing evolution and adaptation. This adaptability is vital for any agent destined to navigate a world in perpetual change. It underscores the importance of continual learning: the process of assimilating new experiences while retaining previously acquired knowledge \citep{wang2023comprehensive}. 

Moreover, \citet{ishiguro2004should} have shown, both through theory and practical application in robotics, that a close and effective integration of control mechanisms with body dynamics significantly enhances energy efficiency. \citet{ororbia2024} elaborates on energy efficiency, proposing that embodied mortal systems, which are characterized by their inherent lifecycle and eventual mortality, can optimize energy usage through adaptive processes. These processes allow the system to self-organize and maintain homeostasis by minimizing free energy, in alignment with the principles of the free energy principle \citep{Friston2010, friston2023free}. Coupled systems lead also to the emergence of intriguing behaviors that can be hard to explicitly program or learn from disembodied datasets \cite{rosas2020reconciling}, an observation aligning with the principles of the TAME framework.

Embodiment is also a prerequisite for learning about \emph{affordances} \citep{Gibson1979}. Learning, or more precisely \emph{realizing} affordances, according to \citet{Vervaeke2012}'s perspectival learning, is a fundamental capacity of \gls{agi}, as affordances are what ``fill our world with meaning'' \citep{roli2022organisms}, and are thus necessary for agents that give meaning to their own world. Affordances emerge from the dynamic interplay between an agent's perception, objectives, abilities, and the characteristics of objects and contexts within the environment; for example, a chair affords us to sit, a glass to drink and a hand to grasp and pick up objects. \citet{roli2022organisms} argue that the capacity to comprehend, utilize, and be influenced by environmental affordances distinguishes biological intelligence from current artificial systems. Besides affordances, \gls{eai} is also indispensable for investigating emergent phenomena such as qualia \cite{locke1847essay, korth2022purpose}, consciousness \cite{Solms2019}, as well as creativity, empathy \citep{Perez2023}, and ethical understanding \citep{lake2017building, russell2021human}.

Finally, there is the important question of why an intelligent agent would do anything in the first place \cite{pfeifer2004embodied}. What drives it to engage and acquire new knowledge without external prompts?
Within well-framed small worlds, such as a chess game, an agent's purpose is straightforward: deciding the next move. However, when navigating large, open worlds, the motivations guiding an agent's decisions grow increasingly ambiguous.
The concepts of active inference and the free energy principle \citep{Friston2010, friston2023free} provide a compelling framework for understanding the behaviors of intelligent agents. This principle posits that minimizing surprise and uncertainty is the core objective of the agents. They achieve this through the use of internal models to forecast outcomes, continually updating these models with sensory input, and proactively modifying their surroundings to better match their expectations. This concept resonates within the AI community, particularly in the design of agents equipped with mechanisms for \emph{intrinsic motivation} \citep{oudeyer2007intrinsic, pathak2017curiosity}, which incentivize agents to explore and acquire new knowledge to reduce uncertainty.

However, what propels an intelligent agent to act, especially beyond mere survival instincts, continues to be a matter of debate. We argue that exploring and developing embodied agents will illuminate this question. Thus, \gls{eai} not only shows potential for significant breakthroughs toward achieving AGI, but also has deep implications for our understanding of cognition in general.
\vspace{-4pt}

\section{Theoretical framework}
\label{sec:theoretical_framework}
\vspace{-4pt}

In previous sections, we have underscored the pivotal role of \gls{eai} in advancing toward AGI. Shifting focus, we now delve into the essential components that, we believe, will comprise \glspl{eai}. We draw heavily on the concept of \emph{cognitive architectures} designed by cognitive scientists aiming to model the human mind \citep{thagard2012cognitive}. 
Despite the promise these architectures hold for enhancing modern machine learning methods, progress on this has been notably limited \citep{Kotseruba2020}. 
The slow advancement is largely due to cognitive architectures being the domain of neuroscientists and cognitive scientists, with only a select few within the machine learning community exploring their potential for AGI. We advocate for a synergistic strategy that marries cognitive architectures with machine learning within the \gls{eai} paradigm, proposing it as a viable path toward AGI. The emergence of agent-based LLMs, such as AutoGPT \citep{firat2023if}, which pioneers the generation of autonomous agents, and PanguAgent \citep{christianos2023pangu}, an agent-focused language model, indicate the potential of this approach.

We identify four essential components of an \gls{eai} system: (i) perception: the ability of the agent to sense its environment; (ii) action: the ability to interact with and change its environment; (iii) memory: the capacity to retain past experiences; and (iv) learning: integrating experiences to form new knowledge and abilities. These components are notably aligned with the active inference framework of \citet{Friston2010}. In this framework, the agent models its world through a probabilistic generative model that infers the causes of its sensory observations (perception). This model is hierarchical, forecasting future states in a top-down manner and reconciling these predictions with bottom-up sensory data, with discrepancies or errors being escalated upwards only when they cannot be reconciled at the initial level. The agent acts to minimize the divergence between its anticipations and reality, thus moving towards states of reduced uncertainty (action). Concurrently, it collects and stores new information about its environment (memory) and refines its internal model to minimize predictive errors (learning). In the sections that follow, we will describe in detail these four components and how they comprise the \gls{eai} agent.

\vspace{-4pt}

\subsection{Perception}
\vspace{-4pt}

At the heart of an embodied agent lies the ability to perceive the world in which it exists. Perception is a process by which raw sensory data is transformed into a structured internal representation, enabling the agent to engage in cognitive tasks. 
The range of inputs that inform perception is vast, encompassing familiar human senses such as vision, hearing, smell, touch, and taste.
It extends to any form of stimuli an agent might encounter, be it force sensors in robotics or signal strength indicators in wireless technology. 
The challenge with sensory data is that it is often not immediately actionable. It typically undergoes a process of transformation, a task where recent advances in machine learning can prove invaluable. The field has seen the development of sophisticated methods for learning feature and embedding spaces, facilitating the conversion of raw data into meaningful information \citep{golinko2019generalized, sivaraman2022emblaze}. A particularly effective strategy has been self-supervised learning to learn such representations. Although much of the research has concentrated on single modalities, such as vision \citep{oquab2023dinov2}, the principles underlying these techniques are universally applicable across different sensory inputs \citep{orhan2022don, lee2019making}. 

\vspace{-4pt}

\subsection{Action}
\vspace{-4pt}

Embodied agents navigate the world by taking actions and observing the outcomes. Acting can be broken down into two  steps: (i) choosing \emph{what} action to undertake next, like deciding to relocate to a specific spot, and (ii) determining \emph{how} to execute this action, such as plotting the course to that location. Actions can further be categorized into \emph{reactive} and \emph{goal-directed} types. 
Reactive actions, akin to human reflexes, occur almost instantaneously in response to stimuli and play a crucial role in an agent's immediate self-preservation by maintaining stability. Goal-directed actions, on the other hand, involve strategic planning and are motivated by high-level objectives. 
Reactive actions are important for self-preservation, with model-free reinforcement learning methods playing an important role for developing reactive control policies in tasks like robot walking \cite{rudin2022learning}. 
On the other hand, for an agent to achieve more complex, high-level objectives, planning is indispensable, even if efficient planning remains an open area of research \citep{lin2022model, shi2022skill}. Central to the concept of planning is the presence of a ``world model'' within the agent, which it can use to predict the consequences of its own actions. 
Model-based RL has made significant strides in developing algorithms that learn these world models and use them for planning \citep{silver2016mastering, kegl2021model, paolo2022guided}.
\vspace{-6pt}

\subsection{Memory}
\vspace{-4pt}

Embodied agents learn from their experience, which are stored in memory. Memory encompasses various dimensions, including its duration (short-term or long-term) and its nature (procedural, declarative, semantic, and episodic). Importantly, memory is not necessarily represented as explicit propositional knowledge; it can be implicitly encoded into the weights of a \gls{nn}.
To navigate cognitive tasks, agents require diverse types of memory systems, each playing a distinct role. Working and short-term memory offer temporary storage to support the agent's immediate objectives. Long-term and episodic memories provide a reservoir for information over longer time. Episodic memory captures and stores unique, perspectival experiences, ready to be accessed when familiar scenarios unfold. Long-term memory, conversely, is the repository for broader propositional knowledge. LLMs, for example, implement long-term memory using \gls{rag} \cite{gao2024retrievalaugmented}, a technique that reduces hallucinations using an external database. This technique showcases how sophisticated machine learning methods can be synergized with cognitive architectures.

\vspace{-4pt}

\subsection{Learning}
\vspace{-4pt}

A defining trait of intelligent agents is their ability to learn.
Yet, how to learn, especially in a continuous and dynamic way,
remains a subject of ongoing research and debate \citep{wang2023comprehensive, yifan2023continual}. While recent strides in AI have largely been powered by training on static datasets, the concept of continual learning, essential for adapting over time, faces challenges. These challenges stem primarily from the inherent limitations of deep NNs, such as catastrophic forgetting \citep{kemker2018measuring}, and the complexities associated with learning from non-stationary data that result from an agent's interaction with its environment \citep{fahrbach2023learning}.
The embodiment hypothesis suggests that true intelligence is born from such interactions \cite{smith2005development}, underscoring the need for dynamic learning methodologies. In this context, simulators emerge as a vital tool, offering a shift away from the static  learning typical of traditional AI. Instead, they enable agents to evolve through ongoing, interactive experiences within simulated environments \citep{duan2022survey}. 

\vspace{-4pt}

\section{Challenges}
\vspace{-4pt}
\label{sec:challenges}

\glspl{eai} agents will adopt an egocentric perspective, experiencing their environment from a first-person viewpoint, in contrast to the allocentric perspective prevalent in current AI systems. This shift is not only essential for meaningful interaction with the world but also offers an advantage by allowing the agents to focus on modeling their immediate surroundings rather than the entirety of the world. 
On the other hand, \glspl{eai} introduces several challenges, including extending current learning theories, managing noise in perception and action effectively and safely, and ensuring meaningful communication with humans that adheres to ethical standards. The remainder of this section will cover these challenges, exploring potential pathways and solutions.

\vspace{-6pt}

\subsection{New learning theory}
\label{sec:new_learning_theory}
\vspace{-4pt}

The principles of \gls{eai} challenge us to reevaluate traditional learning theories \citep{Devroye1996,Vapnik1998}, bridging a gap between supervised and reinforcement learning. Supervised learning, while foundational in AI, assumes that the data is drawn from an unknown but fixed distribution, collected independently of the learning process. This theory gives rise to the classical notions of generalization, over- and underfitting, bias and variance, and asymptotic or finite-sample statistical consistency. This framing is obviously highly useful: even those who are not explicitly doing theory use it transparently as their lingua technica and cognitive scaffolding when working with algorithms and analyzing results.

When embodied agents interact dynamically with their environment, data collection becomes part of the data science pipeline \citep{pfeifer2004embodied, Thrun2005}. Classical supervised learning theory is insufficient to analyze these cases and to guide algorithm building. Extensions, like transfer learning \citep{Pan2010}, multitask learning \citep{Caruana1997}, distribution shift \citep{Quinonero2009}, domain adaptation \citep{Csurka2017} or out-of-distribution generalization, have been proposed to patch basic supervised learning theory, but most of these cling to the original framing, pretending that the data is coming from outside the learning process, encapsulating the value (business or otherwise) of the predictive pipeline. Practically, this is obviously not the case: the data on which we learn a predictor is often collected by the data scientist, responsible for the quality of the pipeline \citep{ONeil2013, Provost2013}. Furthermore, most of the debates around responsible AI turn around the data, not the learning algorithm \citep{ONeil2016, Selbst2019}. Collecting, selecting, and curating data is obviously part of the pipeline. The text we use to train LLMs is created by its writers, rather than drawn from a distribution. In some cases, when collection and model-retraining are automated, the situation may be even worse. For example, in click-through-rate prediction \citep{Bottou2013, Perlich2014} or recommendation systems \citep{deldjoo2020adversarial}, the deployed predictor affects the data for the next round of training, generating an often adversarial feedback. A similar phenomenon is happening in the LLM world: as these AIs become the go-to tools for creative and business writing, the data collected for the next round of training will, in large part, be coming from the previous generation of LLMs.

Reinforcement Learning \citep{Sutton2018} and related paradigms (Bayesian optimization \citep{Mockus1989} or contextual bandits \citep{Langford2008}) offer a closer fit for embodied AI, when the prediction is not the end-product, rather part of a predictive pipeline that also includes data collection. RL affords the data scientist to design a higher-level objective, letting the algorithm optimize both the predictor and the data it is trained on. Here, the mismatch between theory and practice is different from supervised learning. The analysis in RL or bandit theory often focuses on the convergence of the agent to a theoretical optimum, given a fixed but often unknown environment. RL theory usually does not offer tools to analyze the data collected during the learning process, especially when the collection is semi-automatic (includes a human curator in the loop). RL agents, in practice, usually do not converge even in a stationary environment, they rather individuate, making, for example, quite perversely, the random seed part of the algorithm \citep{Henderson2018}. This is even more pronounced in non-stationary environments where the agent's actions alter the environment; a situation which AGI will definitely find itself \cite{DaSilva2006, zhou2024hazard}.

A new learning theory for embodied AI must transcend these limitations. It should account for the dynamic, interactive nature of data in \gls{eai}, where the agent's actions continuously reshape its learning environment. This theory should not just aim for optimal performance in a fixed setting but should embrace a spectrum of behaviors suitable for evolving environments. Moreover, it should provide diagnostics to assess the quality and relevance of data generated through these interactions.
\vspace{-4pt}

\subsection{Noise and uncertainty}
\vspace{-4pt}
\label{noise_uncertainty}
\gls{eai} agents are tasked with navigating the real world, rife with noise and uncertainty. These elements can drastically affect both the agent's perception of its surroundings and the quality of its decision-making. For example, elevated noise levels may distort the agent's interpretation of environmental cues, leading to suboptimal decisions. This challenge is accentuated in an egocentric perspective, where agents frequently encounter continuous streams of fluctuating and imprecise data. Sources of noise include the natural imprecision of sensors and actuators, which might lack accuracy due to manufacturing inconsistencies, degradation over time, or external disturbances. Additionally, quantization error, a byproduct of converting analog signals into digital form \citep{widrow2008quantization}, can further compromise data integrity.

As these agents learn and adapt to their environment, they must also grapple with uncertainty. This uncertainty can obscure the agent's understanding of its environment, influencing its performance. This dilemma is especially prevalent in RL scenarios dealing with partial observability, where decisions must be made with incomplete information, leading to uncertainty in predicting the outcomes of its actions \citep{dulac2021challenges, Hess2023, Pattanaik2017}. 
Therefore, managing noise and uncertainty effectively is paramount for the progress of \gls{eai}.

\vspace{-6pt}

\subsection{Simulators}
\vspace{-4pt}

As we pivot towards \gls{eai}, simulators will assume a fundamental role as a key driver of progress, similar to the role data sets play in the training of traditional \gls{iai} models. These simulators offer a controlled, replicable environment where AI systems can be rigorously trained and tested. This setup allows for learning and adapting to diverse scenarios prior to deployment, ensuring both safety and cost-efficiency. 
A notable advantage of simulators, and requirements, is their speed and ease of parallelization, significantly accelerating training time, making it more feasible to train sophisticated AI models on multiple scenarios simultaneously.

Many advanced simulators have been introduced recently, yet they often demand significant computational resources and are predominantly geared towards robotics applications \citep{li2021igibson, gan2020threedworld, yan2018chalet, puig2018virtualhome, gao2019vrkitchen}. For these simulators to truly serve the needs of \gls{eai}, they must expand their scope to a broader spectrum of environments. A major challenge in the use of simulators is bridging the ``reality gap'' \citep{bousmalis2017closing}: the difference between simulated conditions and the agent's eventual real-world or virtual deployment context \citep{ligot2020simulation}. This gap can lead to a situation where models that excel in simulations fail in actual application, undermining the effectiveness of the training process. Despite numerous strategies being put forward to mitigate the reality gap \cite{salvato2021crossing, daza2023sim, daoudi2023trust, koos2012transferability, tobin2017domain}, it remains an unresolved issue in the field, challenging the applicability of simulated training environments.

\vspace{-4pt}

\subsection{Interaction with humans}
\vspace{-4pt}

A key ambition of \gls{eai} is to seamlessly interact with and learn from humans, enhancing AI's ability to offer personalized and impactful solutions. 
By improving these interactions, \gls{eai} will also diminish fear and mistrust towards AI technologies, leading to broader acceptance and integration. 
In this endeavor, LLMs stand out as particularly beneficial, with their ability to comprehend and produce human-like text, facilitating communication in natural language and making engagements with AI more natural and accessible. 
The domain of Human-Robot Interaction (HRI) offers valuable lessons for enhancing AI-human communication, as researchers in this domain have dedicated efforts to explore innovative methods for robots to better communicate with us \cite{amirova202110, bonarini2020communication}. 
Yet, the challenge of ensuring proper and ethical communication with AI systems persists.
The effectiveness of LLMs, for instance, hinges significantly on their training and how well they are aligned with human intentions and values \citep{wang2023aligning}. 
Integrating human oversight directly into the AI development process and establishing comprehensive guidelines and protocols for AI communication are among the proposed strategies to address these challenges, aiming to make AI interactions more meaningful and ethically sound.

\subsection{Generalization}
An important issue in AI is generalization. 
There have been many attempts at developing systems capable of quickly generalizing to settings unseen at training time \citep{pourpanah2022review} in the same fashion living beings do.
Nonetheless this is still an open problem that will likely afflict embodied AIs as well, as acting in the real world exposes the agent to situations unseen at training time.
For instance, consider a service robot trained in a simulated environment. 
When placed in a real household, it may encounter novel objects and behaviors not present in its training data, leading to suboptimal or even erroneous actions.
This illustrates the critical need for AIs that can adapt and generalize beyond their initial programming.
A promising direction in addressing this problem is the leveraging of the enormous amount of internet data. 
LLMs have demonstrated remarkable zero-shot learning capabilities with minimal fine-tuning \citep{wei2021finetuned}. 
We can envision that some form of pretraining on internet datasets can kick-start the AI before its embodied phase, enhancing generalization and adaptability.

Recent developments in robotics have started exploring this research direction.
\citet{ahn2024autort} used a mixed approach between I-AI and E-AI to effectively control multiple robots in different settings. 
However, only relying on internet data is insufficient.
An important aspect is also the ability to accurately identify unknown situations and avoid overconfidence, a common shortcoming of LLMs, that often produce plausible-sounding response that are factually incorrect \cite{xiong2024llms}. 
The ability to assess its own uncertainty is essential, and can prompt the AI to seek human assistance, similarly to how infants ask for help in their early development.
We believe that, while the integration of I-AI and E-AI will prove necessary as foundation for the development of the next generation of intelligent systems, the active learning paradigm and precise uncertainty estimation are vital.
Active learning, where the AI actively queries for information when uncertain, combined with reliable uncertainty estimation, can enable an E-AI agent to manage novel situations effectively.

Finally, we believe that to properly address the issue of generalization, the community must first clearly define what is the meaning of ``generalization''.
Currently, discussions around this issue often rely on vague terms, referring to an agent's ability to adapt to unseen settings or data. 
However, without a formal definition, it is challenging to assess or improve generalization effectively.

Consider the varying degrees of generalization required in different scenarios: transferring skills from driving a car to driving a bus represents generalization within a similar domain, whereas adapting from walking to swimming involves a more profound shift in the type of task. 
These examples illustrate the spectrum of generalization challenges that embodied AI might face.

To advance this field, it is imperative to develop a precise definition of generalization and establish standardized metrics and benchmarks for measuring an AI's generalization capabilities \cite{kawaguchi2017generalization}.
This necessity ties back to our discussion in Section~\ref{sec:new_learning_theory}, highlighting the urgent need for a new learning theory that can provide a principled approach to developing agents that generalize well.
Addressing these questions will hopefully lead to more precise and principled approaches in the development of the field.

\subsection{Hardware limitations}
\vspace{-4pt}

A significant challenge to the broad-scale development and integration of \gls{eai} lies in the hardware requirements of these AI systems. Presently, AI technologies largely depend on GPU clusters, which are, while powerful, not ideally suited for embodied agents due to their high cost, energy consumption, and extensive heat output. Additionally, the physical bulk and heft of GPUs pose logistical challenges for mobile agents or those operating within spatial limitations. Addressing these constraints necessitates the innovation of new, energy-efficient hardware solutions that can be embedded within the agents. Promising developments are on the horizon, with Google's Tensor Processing Unit (TPU) \citep{norrie2021design, cass2019taking} and Huawei's Ascend chip \citep{liao2021ascend} leading the charge. These advancements, coupled with the potential of neuromorphic computing and the strategic synergy of hardware-software co-design, signal a new era of hardware capability. Moreover, the development of energy and data-efficient algorithms is critical. Such breakthroughs in hardware and algorithm efficiency will have a direct and profound effect on an AI's ability to understand, decide, and interact within its environment, enabling \gls{eai} agents to operate more autonomously and effectively in a diverse array of settings.

\vspace{-6pt}

\section{Conclusion}
\label{sec:conclusion}
\vspace{-4pt}

In this paper, we have articulated the critical role Embodied AI plays on the path toward achieving AGI, setting it apart from prevailing AI methodologies, notably LLMs. By integrating insights from a spectrum of research fields, we underscored how E-AI's development benefits from existing knowledge, with LLMs enhancing the potential for intuitive interactions between humans and emerging AI entities. We introduced a comprehensive theoretical framework for the development of E-AI, grounded in the principles of cognitive science, highlighting perception, action, memory, and learning, situating E-AI within the context of Friston’s active inference framework, thereby offering a wide-ranging theoretical backdrop for our discussion. Despite the outlook, the journey ahead is fraught with challenges, not least the formulation of a novel learning theory tailored for AI and the creation of sophisticated hardware solutions. This paper aims to serve as a roadmap for ongoing and future research into E-AI, proposing directions that could lead to significant advancements in the field.

\section*{Impact Statement}

While the development of Embodied AI introduces complexities and challenges, particularly in hardware requirements, ethical considerations, and safety protocols, the potential benefits significantly outweigh these drawbacks. E-AI stands to evolve our interaction with technology, imbuing AI with a deeper understanding of and engagement with both the physical world and human society. This not only paves the way for more natural and effective human-AI interactions but also enhances AI's adaptability and application across a broad spectrum of fields.

\bibliography{main}
\bibliographystyle{icml2024}

\end{document}